\documentclass[conference]{IEEEtran}
\IEEEoverridecommandlockouts
\usepackage{cite}
\usepackage{amsmath,amssymb,amsfonts}
\usepackage{algorithmic}
\usepackage{graphicx}
\usepackage{textcomp}
\usepackage{xcolor}

\def\BibTeX{{\rm B\kern-.05em{\sc i\kern-.025em b}\kern-.08em
    T\kern-.1667em\lower.7ex\hbox{E}\kern-.125emX}}

\usepackage{titlesec}
\titlespacing*{\section}{0pt}{1.0ex}{0.8ex}
\titlespacing*{\subsection}{0pt}{0.8ex}{0.6ex}
\makeatletter
\newcommand{\linebreakand}{%
  \end{@IEEEauthorhalign}
  \hfill\mbox{}\par
  \mbox{}\hfill\begin{@IEEEauthorhalign}
}
\makeatother

\begin{document}

\title{Hypergraph and Latent ODE Learning for Multimodal Root Cause Localization in Microservices\\}

\author{
\IEEEauthorblockN{Xin Liu*}
\IEEEauthorblockA{\textit{Independent Researcher} \\
New York, USA \\
iamxinliu@gmail.com}
\and
\IEEEauthorblockN{Yuhang He}
\IEEEauthorblockA{\textit{Independent Researcher} \\
Chicago, USA \\
yuhang.he@outlook.com}
\and
\IEEEauthorblockN{Sichen Zhao}
\IEEEauthorblockA{\textit{Northeastern University} \\
Boston, USA \\
zhao.siche@northeastern.edu}
\and
\linebreakand
\IEEEauthorblockN{Kejian Tong}
\IEEEauthorblockA{\textit{Independent researcher} \\
 Mukilteo, USA \\
tongcs2021@gmail.com}
\and
\IEEEauthorblockN{Xingyu Zhang}
\IEEEauthorblockA{\textit{Independent researcher} \\
 Texas, USA \\
xingyu\_zhang@gwmail.gwu.edu}
}

\maketitle

\begin{abstract}
Root cause localization in cloud native microservice systems requires modeling complex service dependencies, irregular temporal dynamics, and heterogeneous observability data. We present HyperODE RCA, a unified framework that combines hypergraph attention learning, latent ordinary differential equations, and multimodal cross attention fusion for fine grained root cause analysis. The method learns higher order service interactions through differentiable hyperedge construction, captures continuous anomaly evolution from irregular observations with an ODE RNN encoder, and adaptively fuses logs, traces, metrics, entities, and events using context aware modality routing. We further improve robustness with a variational information bottleneck, temporal causal regularization, and invariant risk constraints. Experiments on the Tianchi AIOps benchmark show clear gains over strong baselines in ranking and classification performance, while preserving interpretability through learned hypergraph attention.
\end{abstract}


\section{Introduction}
Cloud native applications are commonly deployed as large microservice systems, where failures propagate across loosely coupled components and produce heterogeneous evidence in logs, traces, metrics, and events. Advances in continuous-time modeling, higher-order relational learning, and multimodal representation learning create new opportunities for more faithful diagnosis of such systems \cite{kidger2020neural,chien2021you,nagrani2021attention}.

Existing root cause localization pipelines still have three major limitations. First, pairwise service graphs can overlook coordinated failures involving multiple services. Second, discrete sequence models handle irregular observations poorly and do not explicitly capture propagation speed. Third, simple fusion strategies underuse complementary observability signals, reducing causal attribution under noisy incidents \cite{kidger2020neural,chien2021you}.

We propose HyperODE RCA, a unified framework for multimodal root cause localization in microservice environments. It learns data-driven hypergraphs for higher-order service interactions, uses a latent ODE encoder to model continuous anomaly evolution, and applies cross-attention fusion with adaptive modality routing. Temporal causal regularization and a variational bottleneck with invariant constraints further improve diagnostic accuracy and interpretability across diverse fault scenarios \cite{nagrani2021attention}.

\section{Related Work}
Early lines of recent work on system diagnosis increasingly connect root cause analysis with temporal dependency discovery and causal structure learning. Xue et al.\cite{xue2026resilient} proposed a risk-aware dynamic routing framework that combines a GCN-GRU spatiotemporal graph model with dynamic edge-weight adjustment, enabling prediction-guided adaptation to time-varying network risk. For HyperODE-RCA, this idea is technically relevant because it suggests modeling service dependencies as dynamically reweighted graph relations conditioned on predicted anomaly evolution, which can strengthen fault-propagation reasoning in microservice systems. Methods for learning directed relations from time series provide a useful foundation because they formalize how events in one component may precede changes in another, yet they typically assume relatively simple variable interactions and are not designed for heterogeneous observability streams in microservice systems \cite{pamfil2020dynotears} \cite{brouillard2020differentiable}.Wang et al.\cite{wang2025self} introduced SEAM, a defense that couples optimization on benign and harmful data, together with adversarial gradient ascent, so that models retain legitimate capabilities while becoming intrinsically resistant to harmful fine-tuning. This robustness-oriented training perspective is relevant to HyperODE-RCA because it motivates objectives that preserve diagnostic utility while reducing brittle adaptation to spurious or adversarial observability patterns.

A second line of work studies stronger representation learning and fusion architectures for complex inputs. Gao et al.\cite{gao2024leveraging} introduced an enhanced retrieval-augmented generation framework that couples ScaNN-based nearest-neighbor search with the Gemma language model, improving retrieval efficiency, scalability, and context-aware response generation. For our setting, this line of work suggests a practical way to strengthen the retrieval and contextual grounding of heterogeneous observability evidence before multimodal fusion, thereby complementing HyperODE-RCA’s downstream diagnosis pipeline. Multimodal transformers improve information exchange across heterogeneous signals, while scalable latent architectures support flexible processing of structured inputs. These advances motivate RCA models that move beyond handcrafted aggregation and allow incident dependent weighting of different evidence sources\cite{jaegle2021perceiver}. Tong et al.\cite{tong2024integrated} proposed an integrated machine-learning and deep-learning framework that combines neural networks with ensemble base learners and uses SMOTE-based preprocessing to improve robustness under class imbalance. For HyperODE-RCA, this hybrid classification strategy is a relevant complement to the final root-cause scoring stage, suggesting a practical way to improve decision reliability when fault classes are unevenly distributed.

Our method is also closely related to research on irregular time series modeling and hypergraph learning. Neural controlled differential models show the value of continuous dynamics for uneven observations, and recent hypergraph networks demonstrate how higher order relations can be learned more effectively than pairwise graphs alone. In addition, information bottleneck based robust learning offers a principled way to reduce spurious correlations during prediction \cite{garg2015learning} \cite{alemi2016deep}. Wang et al.\cite{wang2025reasoning} showed that large reasoning models can rely on retrieval shortcuts that compete with explicit reasoning, and introduced FARL to suppress such shortcut behavior through memory unlearning combined with reinforcement learning. For our setting, this perspective further motivates mechanisms that reduce reliance on spurious observability cues and strengthen diagnosis based on causally informative multimodal evidence. Long et al.\cite{long2024enhancing}  proposed a Transformer-based matching framework trained with InfoNCE loss and knowledge distillation, which improves robustness to noisy related samples and increases matching accuracy and efficiency. In our setting, these ideas can strengthen noise-resistant alignment among heterogeneous observability signals, particularly by complementing contrastive log template clustering and cross-modal representation learning.

\section{Methodology}
\begin{figure}[htbp]
\centering
\includegraphics[width=0.5\textwidth]{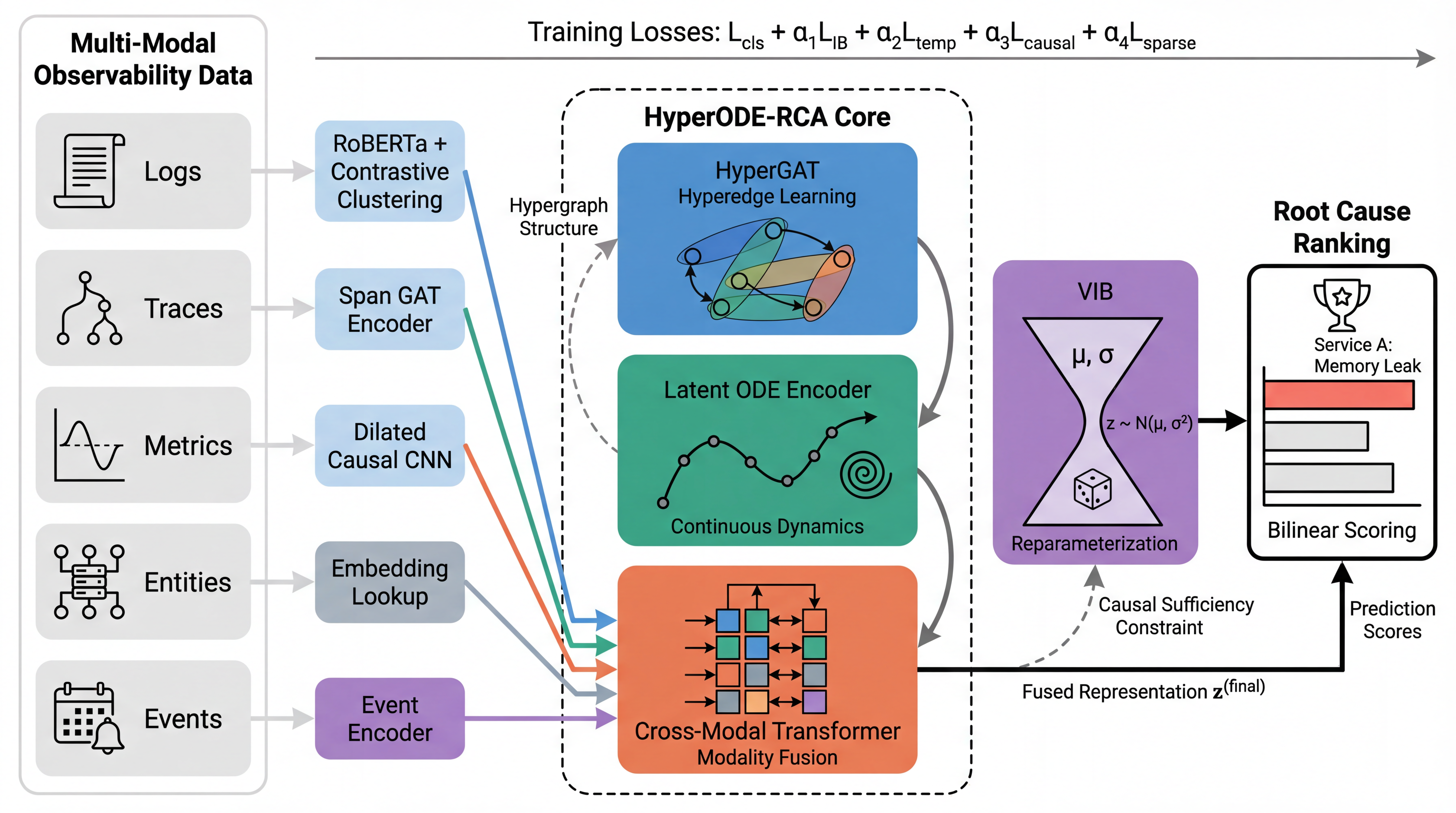}
\caption{Multi-modal observability signals (logs, traces, metrics, entities, events) are encoded by modality-specific encoders, then processed through three core modules.}
\label{fig:148_1}
\end{figure}

We propose HyperODE-RCA for root cause localization in cloud-native microservices. Unlike prior methods based on pairwise dependencies and discrete-time assumptions, it models higher-order fault propagation and continuous anomaly evolution. The framework combines HyperGAT, which learns multi-service dependencies through differentiable hyperedge generation with Gumbel-Softmax relaxation, and a Latent ODE encoder, which captures continuous-time dynamics, supports interpolation at arbitrary times, and derives anomaly propagation velocity. Logs, traces, metrics, and events are fused by Transformer-based cross-attention with learned modality routing, where asymmetric inter-modal attention improves causal inference. The model further incorporates a variational information bottleneck, invariant-risk-minimization-based causal sufficiency constraints, temporal causal masking, contrastive log template clustering, and median-absolute-deviation normalization. On the Tianchi AIOps benchmark, HyperODE-RCA outperforms MicroRCA, CloudRanger, and recent causal discovery methods, while providing interpretable explanations through attention visualization over the learned hypergraph. The overall architecture is shown in Fig.~\ref{fig:148_1}.

\section{Algorithm and Model}

\subsection{HyperGAT: Hypergraph Attention Network}

\begin{figure}[htbp]
\centering
\includegraphics[width=0.5\textwidth]{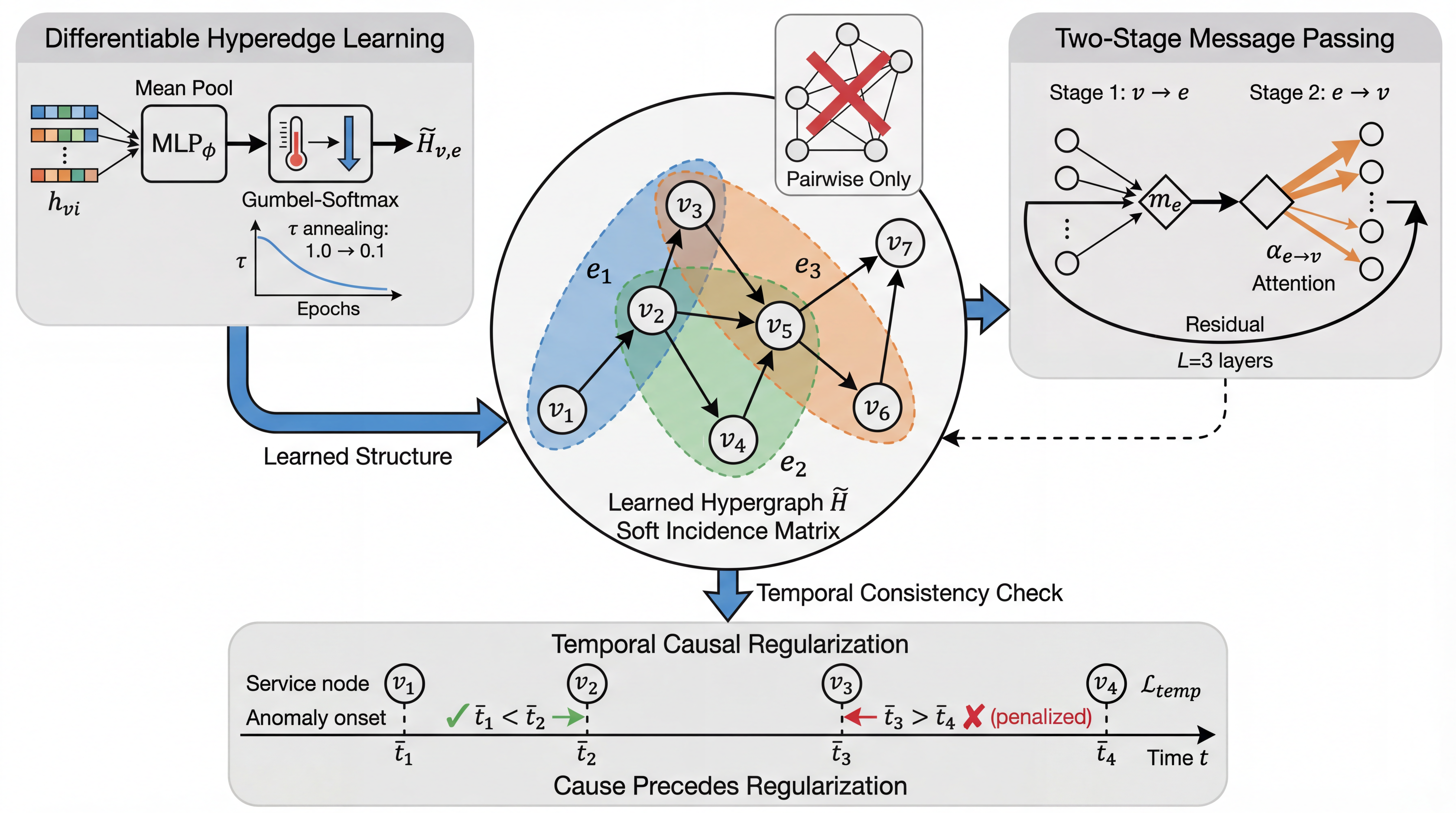}
\caption{HyperGAT module. (Center) Learned hypergraph with overlapping hyperedges capturing multi-service failure patterns. (Top-left) Differentiable hyperedge learning via Gumbel-Softmax with temperature annealing. (Top-right) Two-stage message passing: vertex-to-hyperedge aggregation followed by attention-weighted hyperedge-to-vertex distribution. (Bottom) Temporal causal regularization penalizing backward-in-time attention weights.}
\label{fig:148_2}
\end{figure}

Pairwise graph models are insufficient for microservice failures caused by coordinated multi-service interactions. HyperGAT addresses this with learned hyperedges and causal message passing, as illustrated in Fig.~\ref{fig:148_2}.

\subsubsection{Hypergraph Representation}

We model the service topology as a hypergraph $\mathcal{H} = (\mathcal{V}, \mathcal{E}_h)$ with incidence matrix $\mathbf{H} \in \{0,1\}^{|\mathcal{V}| \times |\mathcal{E}_h|}$:
\begin{equation}
H_{v,e} = \begin{cases}
1 & \text{if } v \in e \\
0 & \text{otherwise}
\end{cases}
\end{equation}
Hyperedges are inferred from data rather than predefined.

\subsubsection{Differentiable Hyperedge Learning}

To enable gradient-based learning, hyperedge assignment is relaxed by Gumbel-Softmax. For candidate hyperedge $e_k$ over service subset $S_k$,
\begin{equation}
\ell_k = \mathbf{w}_e^T \cdot \text{MLP}_\phi\left(\frac{1}{|S_k|}\sum_{v_i \in S_k} \mathbf{h}_{v_i}\right)
\end{equation}
and the soft incidence value is
\begin{equation}
\tilde{H}_{v,e_k} = \frac{\exp((\ell_k + g_1)/\tau)}{\exp((\ell_k + g_1)/\tau) + \exp((- \ell_k + g_2)/\tau)}
\end{equation}
where $g_1, g_2 \sim \text{Gumbel}(0,1)$ and $\tau > 0$. We anneal $\tau$ from 1.0 to 0.1 and restrict candidate hyperedges using trace call graphs and log co-occurrence.

\subsubsection{Hypergraph Convolution with Causal Attention}

With $\tilde{\mathbf{H}}$, HyperGAT performs two-stage message passing. Degree matrices are
\begin{equation}
\tilde{D}_v = \text{diag}\left(\sum_{e} \tilde{H}_{v,e}\right), \quad \tilde{D}_e = \text{diag}\left(\sum_{v} \tilde{H}_{v,e}\right)
\end{equation}
Hyperedge messages are
\begin{equation}
\mathbf{m}_e = \sum_{u \in e} \frac{\tilde{H}_{u,e}}{\tilde{D}_e[e,e]} \cdot \mathbf{W}_v \mathbf{h}_u
\end{equation}
and vertex updates are
\begin{equation}
\mathbf{h}_v^{(l+1)} = \sigma\left(\sum_{e \ni v} \alpha_{e \rightarrow v} \cdot \mathbf{W}_e \mathbf{m}_e\right)
\end{equation}
with attention
\begin{equation}
\alpha_{e \rightarrow v} = \frac{\exp\left(\text{LeakyReLU}\left(\mathbf{a}^T[\mathbf{W}_q\mathbf{h}_v \| \mathbf{m}_e]\right)\right)}{\sum_{e' \ni v} \exp\left(\text{LeakyReLU}\left(\mathbf{a}^T[\mathbf{W}_q\mathbf{h}_v \| \mathbf{m}_{e'}]\right)\right)}
\end{equation}
We use $L=3$ layers with residual connections.

\subsubsection{Temporal Causal Regularization}

Temporal precedence is encouraged by penalizing backward attention:
\begin{equation}
\mathcal{L}_{\text{temp}} = \sum_{e \in \mathcal{E}_h} \sum_{(u,v) \in e \times e} \tilde{H}_{u,e} \cdot \tilde{H}_{v,e} \cdot \alpha_{u \rightarrow v} \cdot \max(0, \bar{t}_u - \bar{t}_v)
\end{equation}
where $\bar{t}_v$ is the mean anomaly onset time estimated from metric deviations.

\subsection{Latent ODE Encoder for Continuous Dynamics}

\begin{figure}[htbp]
\centering
\includegraphics[width=0.5\textwidth]{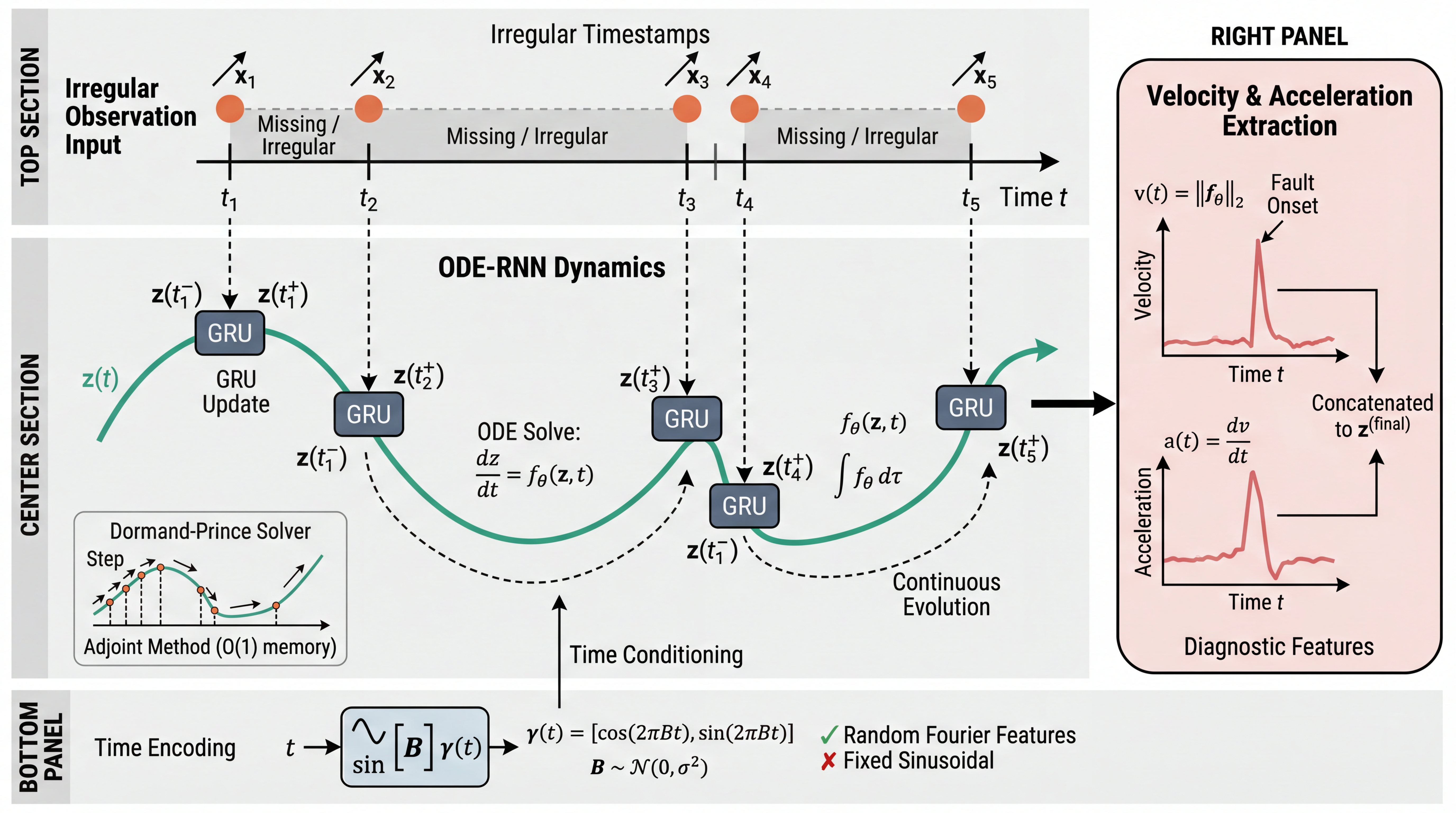}
\caption{Latent ODE Encoder. Irregularly sampled observations (orange dots) are integrated into a continuous latent trajectory $\mathbf{z}(t)$ via neural ODE dynamics. At each observation time, a GRU cell updates the state by incorporating new evidence. The continuous formulation enables extraction of propagation velocity $v(t)$ and acceleration $a(t)$ as diagnostic features, which spike at fault onset.}
\label{fig:148_3}
\end{figure}

Because observability signals are irregularly sampled while fault propagation is continuous, we model latent dynamics with a neural ODE and ODE-RNN encoder, shown in Fig.~\ref{fig:148_3}.

\subsubsection{Neural ODE Formulation}

The latent state $\mathbf{z}(t) \in \mathbb{R}^d$ evolves as
\begin{equation}
\frac{d\mathbf{z}(t)}{dt} = f_\theta(\mathbf{z}(t), t)
\end{equation}
with vector field
\begin{equation}
f_\theta(\mathbf{z}, t) = \mathbf{W}_2 \cdot \text{SiLU}(\mathbf{W}_1[\mathbf{z} \| \gamma(t)] + \mathbf{b}_1) + \mathbf{b}_2
\end{equation}
and time encoding
\begin{equation}
\gamma(t) = [\cos(2\pi \mathbf{B} t), \sin(2\pi \mathbf{B} t)]
\end{equation}
where $\mathbf{B} \in \mathbb{R}^{d_\gamma/2}$ is fixed after sampling.

\subsubsection{ODE-RNN for Irregular Observations}

For observations $\{(t_i, \mathbf{x}_i)\}_{i=1}^N$, latent dynamics between timestamps follow
\begin{equation}
\mathbf{z}(t_i^-) = \mathbf{z}(t_{i-1}^+) + \int_{t_{i-1}}^{t_i} f_\theta(\mathbf{z}(\tau), \tau) d\tau
\end{equation}
solved by a Dormand-Prince adaptive solver. At each observation, a GRU updates the state:
\begin{equation}
\mathbf{r}_i = \sigma(\mathbf{W}_{xr}\mathbf{x}_i + \mathbf{W}_{hr}\mathbf{z}(t_i^-) + \mathbf{b}_r)
\end{equation}
\begin{equation}
\mathbf{u}_i = \sigma(\mathbf{W}_{xu}\mathbf{x}_i + \mathbf{W}_{hu}\mathbf{z}(t_i^-) + \mathbf{b}_u)
\end{equation}
\begin{equation}
\tilde{\mathbf{z}}_i = \tanh(\mathbf{W}_{xz}\mathbf{x}_i + \mathbf{W}_{hz}(\mathbf{r}_i \odot \mathbf{z}(t_i^-)) + \mathbf{b}_z)
\end{equation}
\begin{equation}
\mathbf{z}(t_i^+) = (1 - \mathbf{u}_i) \odot \mathbf{z}(t_i^-) + \mathbf{u}_i \odot \tilde{\mathbf{z}}_i
\end{equation}
Gradients are computed by the adjoint sensitivity method:
\begin{equation}
\frac{d\mathbf{a}(t)}{dt} = -\mathbf{a}(t)^T \frac{\partial f_\theta(\mathbf{z}(t), t)}{\partial \mathbf{z}}
\end{equation}
where $\mathbf{a}(t) = \partial \mathcal{L} / \partial \mathbf{z}(t)$.

\subsubsection{Propagation Velocity Features}

Continuous dynamics also provide derivative-based diagnostics:
\begin{equation}
v(t) = \left\|f_\theta(\mathbf{z}(t), t)\right\|_2
\end{equation}
\begin{equation}
a(t) = \frac{d v(t)}{dt} = \frac{f_\theta^T \cdot J_{f_\theta} \cdot f_\theta}{\|f_\theta\|_2}
\end{equation}
where $J_{f_\theta} = \partial f_\theta / \partial \mathbf{z}$. Velocity and acceleration are appended to the final representation.

\subsection{Multi-Modal Fusion via Cross-Attention Transformer}

\begin{figure}[htbp]
\centering
\includegraphics[width=0.5\textwidth]{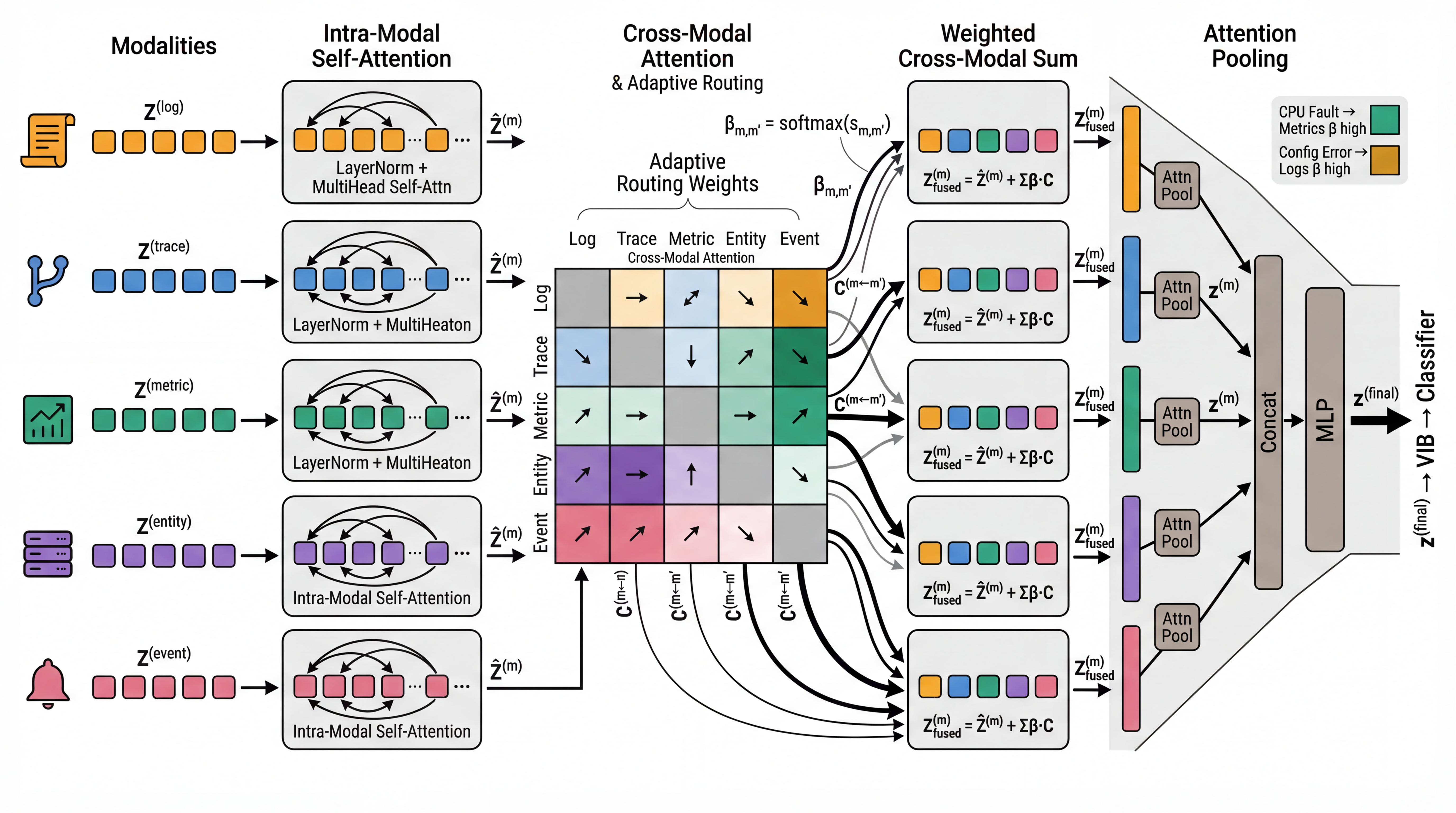}
\caption{Multi-modal cross-attention fusion architecture. Each modality is first refined through intra-modal self-attention, then cross-modal attention with adaptive routing weights $\beta_{m,m'}$ enables modalities to selectively attend to each other based on incident context. Hierarchical attention pooling aggregates token-level representations into a unified embedding $\mathbf{z}^{(\text{final})}$.}
\label{fig:148_4}
\end{figure}

Logs, traces, metrics, entities, and events are first encoded separately and then fused through structured cross-attention with adaptive routing, as shown in Fig.~\ref{fig:148_4}.

\subsubsection{Modality-Specific Encoding}

Logs are encoded by pre-trained RoBERTa after contrastive template clustering:
\begin{equation}
p(j | l_i) = \frac{\exp(\text{sim}(\mathbf{h}_{l_i}, \mathbf{p}_j) / \tau_c)}{\sum_{j'} \exp(\text{sim}(\mathbf{h}_{l_i}, \mathbf{p}_{j'}) / \tau_c)}
\end{equation}
where $\text{sim}(\cdot, \cdot)$ is cosine similarity and $\tau_c = 0.07$.

Trace spans are encoded by a 2-layer GAT:
\begin{equation}
\mathbf{h}_i^{(l+1)} = \sigma\left(\sum_{j \in \mathcal{N}(i)} \alpha_{ij} \mathbf{W}^{(l)} \mathbf{h}_j^{(l)}\right)
\end{equation}

Metric sequences are modeled by dilated causal convolutions:
\begin{equation}
\mathbf{y}^{(l)} = \text{ReLU}(\text{Conv1D}(\mathbf{y}^{(l-1)}; \text{dilation}=2^l))
\end{equation}

\subsubsection{Cross-Modal Transformer}

Given modality token sequences $\{\mathbf{Z}^{(m)}\}_{m \in \mathcal{M}}$, intra-modal self-attention is
\begin{equation}
\hat{\mathbf{Z}}^{(m)} = \text{LayerNorm}(\mathbf{Z}^{(m)} + \text{MultiHead}(\mathbf{Z}^{(m)}, \mathbf{Z}^{(m)}, \mathbf{Z}^{(m)}))
\end{equation}
Cross-modal attention from $m'$ to $m$ is
\begin{equation}
\mathbf{C}^{(m \leftarrow m')} = \text{MultiHead}(\hat{\mathbf{Z}}^{(m)}, \hat{\mathbf{Z}}^{(m')}, \hat{\mathbf{Z}}^{(m')})
\end{equation}
with
\begin{equation}
\text{MultiHead}(\mathbf{Q}, \mathbf{K}, \mathbf{V}) = \text{Concat}(\text{head}_1, \ldots, \text{head}_h)\mathbf{W}^O
\end{equation}
\begin{equation}
\text{head}_i = \text{softmax}\left(\frac{\mathbf{Q}\mathbf{W}_i^Q(\mathbf{K}\mathbf{W}_i^K)^T}{\sqrt{d_k}}\right)\mathbf{V}\mathbf{W}_i^V
\end{equation}
Routing weights depend on incident context:
\begin{equation}
\beta_{m,m'} = \frac{\exp(s_{m,m'})}{\sum_{m''} \exp(s_{m,m''})}, \quad s_{m,m'} = \mathbf{w}^T[\bar{\mathbf{z}}^{(m)} \| \bar{\mathbf{z}}^{(m')} \| \mathbf{a}]
\end{equation}
and fused modality representations are
\begin{equation}
\mathbf{Z}^{(m)}_{\text{fused}} = \hat{\mathbf{Z}}^{(m)} + \sum_{m' \neq m} \beta_{m,m'} \cdot \mathbf{C}^{(m \leftarrow m')}
\end{equation}

\subsubsection{Hierarchical Aggregation}

Each modality is pooled by attention:
\begin{equation}
\mathbf{z}^{(m)} = \sum_{i=1}^{|\mathbf{Z}^{(m)}|} \frac{\exp(\mathbf{w}_p^T \mathbf{z}_i^{(m)})}{\sum_j \exp(\mathbf{w}_p^T \mathbf{z}_j^{(m)})} \mathbf{z}_i^{(m)}
\end{equation}
The final fused representation is
\begin{equation}
\mathbf{z}^{(\text{final})} = \text{MLP}([\mathbf{z}^{(\text{log})} \| \mathbf{z}^{(\text{trace})} \| \mathbf{z}^{(\text{metric})} \| \mathbf{z}^{(\text{entity})} \| \mathbf{z}^{(\text{event})}])
\end{equation}

\subsection{Variational Information Bottleneck with Causal Constraints}

To suppress spurious correlations, we combine information bottleneck learning with invariance constraints.

\subsubsection{Information Bottleneck Principle}

Let $\mathbf{X}$ denote observations, $\mathbf{Z}$ the representation, and $\mathbf{Y}$ the root-cause labels. The IB objective is
\begin{equation}
\min_\theta I(\mathbf{X}; \mathbf{Z}) - \beta I(\mathbf{Z}; \mathbf{Y})
\end{equation}
We optimize a variational form by sampling
\begin{equation}
\mathbf{z} = \boldsymbol{\mu}_\phi(\mathbf{x}) + \boldsymbol{\sigma}_\phi(\mathbf{x}) \odot \boldsymbol{\epsilon}, \quad \boldsymbol{\epsilon} \sim \mathcal{N}(\mathbf{0}, \mathbf{I})
\end{equation}
and using the regularizer
\begin{equation}
D_{\text{KL}} = \frac{1}{2}\sum_{j=1}^d \left(\mu_j^2 + \sigma_j^2 - \log \sigma_j^2 - 1\right)
\end{equation}

\subsubsection{Causal Sufficiency Constraint}

For environments $\mathcal{E} = \{e_1, \ldots, e_E\}$ defined by time periods or service subsets, we enforce invariant performance:
\begin{equation}
\mathcal{L}_{\text{causal}} = \text{Var}_{e \in \mathcal{E}}[\mathcal{L}_e(\theta)] + \lambda_{\text{grad}} \|\nabla_\theta \mathcal{L}_e(\theta)\|_2^2
\end{equation}

\subsection{Root Cause Classification}

Each candidate $c_k = \langle\text{service}_k, \text{fault\_type}_k\rangle$ is embedded as
\begin{equation}
\mathbf{e}_{c_k} = \mathbf{E}_{\text{svc}}[\text{service}_k] + \mathbf{E}_{\text{fault}}[\text{fault\_type}_k]
\end{equation}
and scored by
\begin{equation}
s_k = \mathbf{z}^{(\text{final})T} \mathbf{W}_{\text{bil}} \mathbf{e}_{c_k} + \mathbf{w}_z^T \mathbf{z}^{(\text{final})} + \mathbf{w}_c^T \mathbf{e}_{c_k} + b, \quad \hat{y}_k = \sigma(s_k)
\end{equation}
where $\mathbf{W}_{\text{bil}}$ is factorized as $\mathbf{U}\mathbf{V}^T$ with rank 64.

\subsection{Training Objective}

The full objective is
\begin{equation}
\mathcal{L} = \mathcal{L}_{\text{cls}} + \alpha_1 \mathcal{L}_{\text{IB}} + \alpha_2 \mathcal{L}_{\text{temp}} + \alpha_3 \mathcal{L}_{\text{causal}} + \alpha_4 \mathcal{L}_{\text{sparse}}
\end{equation}
with classification loss
\begin{equation}
\mathcal{L}_{\text{cls}} = -\frac{1}{K}\sum_{k=1}^K [\tilde{y}_k \log \hat{y}_k + (1-\tilde{y}_k)\log(1-\hat{y}_k)]
\end{equation}
where $\tilde{y}_k = (1-\epsilon)y_k + \epsilon/K$. The sparsity term $\mathcal{L}_{\text{sparse}} = \|\tilde{\mathbf{H}}\|_1 / (|\mathcal{V}| \cdot |\mathcal{E}_h|)$ encourages interpretable hypergraphs. Training uses AdamW with learning rate $3 \times 10^{-4}$, weight decay 0.01, and cosine annealing with 10\% warmup.

\subsection{Data Preprocessing}

\subsubsection{Metric Normalization}

Heavy-tailed production metrics are normalized by median absolute deviation:
\begin{equation}
\tilde{m}_{t,k} = \frac{m_{t,k} - \text{median}_k}{1.4826 \cdot \text{MAD}_k + \epsilon}
\end{equation}
where $\text{MAD}_k = \text{median}(|m_{t,k} - \text{median}_k|)$.

\subsubsection{Missing Value Handling}

Short gaps are linearly interpolated, and longer gaps use exponential decay:
\begin{equation}
\hat{m}_t = m_{t_{\text{last}}} \cdot \exp(-\lambda(t - t_{\text{last}}))
\end{equation}
with $\lambda = 0.1$ per minute.

\section{Evaluation Metrics}

We adopt five metrics to comprehensively evaluate model performance.

\subsection{F1-Score}

The primary metric balancing precision and recall:
\begin{equation}
\text{Precision} = \frac{\text{TP}}{\text{TP} + \text{FP}}, \quad \text{Recall} = \frac{\text{TP}}{\text{TP} + \text{FN}}
\end{equation}
\begin{equation}
\text{F1} = \frac{2 \cdot \text{Precision} \cdot \text{Recall}}{\text{Precision} + \text{Recall}} = \frac{2\text{TP}}{2\text{TP} + \text{FP} + \text{FN}}
\end{equation}

\subsection{Mean Reciprocal Rank}

MRR evaluates ranking quality:
\begin{equation}
\text{MRR} = \frac{1}{N}\sum_{i=1}^{N} \frac{1}{\text{rank}_i}
\end{equation}
where $\text{rank}_i$ is the position of the first correct prediction for incident $i$.

\subsection{Matthews Correlation Coefficient}

MCC provides balanced evaluation for imbalanced classification:
\begin{equation}
\text{MCC} = \frac{\text{TP} \cdot \text{TN} - \text{FP} \cdot \text{FN}}{\sqrt{(\text{TP}+\text{FP})(\text{TP}+\text{FN})(\text{TN}+\text{FP})(\text{TN}+\text{FN})}}
\end{equation}

\subsection{Area Under ROC Curve}

AUC measures discrimination capability across all thresholds:
\begin{equation}
\text{AUC} = \int_0^1 \text{TPR}(t) \, d\text{FPR}(t) = P(\hat{y}_{\text{pos}} > \hat{y}_{\text{neg}})
\end{equation}

\bibliographystyle{IEEEtran}
\bibliography{references}

@article{kidger2020neural,
  title={Neural controlled differential equations for irregular time series},
  author={Kidger, Patrick and Morrill, James and Foster, James and Lyons, Terry},
  journal={Advances in neural information processing systems},
  volume={33},
  pages={6696--6707},
  year={2020}
}

@article{chien2021you,
  title={You are allset: A multiset function framework for hypergraph neural networks},
  author={Chien, Eli and Pan, Chao and Peng, Jianhao and Milenkovic, Olgica},
  journal={arXiv preprint arXiv:2106.13264},
  year={2021}
}

@article{nagrani2021attention,
  title={Attention bottlenecks for multimodal fusion},
  author={Nagrani, Arsha and Yang, Shan and Arnab, Anurag and Jansen, Aren and Schmid, Cordelia and Sun, Chen},
  journal={Advances in neural information processing systems},
  volume={34},
  pages={14200--14213},
  year={2021}
}

@inproceedings{pamfil2020dynotears,
  title={Dynotears: Structure learning from time-series data},
  author={Pamfil, Roxana and Sriwattanaworachai, Nisara and Desai, Shaan and Pilgerstorfer, Philip and Georgatzis, Konstantinos and Beaumont, Paul and Aragam, Bryon},
  booktitle={International conference on artificial intelligence and statistics},
  pages={1595--1605},
  year={2020},
  organization={Pmlr}
}

@article{brouillard2020differentiable,
  title={Differentiable causal discovery from interventional data},
  author={Brouillard, Philippe and Lachapelle, S{\'e}bastien and Lacoste, Alexandre and Lacoste-Julien, Simon and Drouin, Alexandre},
  journal={Advances in Neural Information Processing Systems},
  volume={33},
  pages={21865--21877},
  year={2020}
}

@article{jaegle2021perceiver,
  title={Perceiver io: A general architecture for structured inputs \& outputs},
  author={Jaegle, Andrew and Borgeaud, Sebastian and Alayrac, Jean-Baptiste and Doersch, Carl and Ionescu, Catalin and Ding, David and Koppula, Skanda and Zoran, Daniel and Brock, Andrew and Shelhamer, Evan and others},
  journal={arXiv preprint arXiv:2107.14795},
  year={2021}
}

@book{garg2015learning,
  title={Learning Apache Kafka},
  author={Garg, Nishant},
  year={2015},
  publisher={Packt Publishing}
}

@article{alemi2016deep,
  title={Deep variational information bottleneck},
  author={Alemi, Alexander A and Fischer, Ian and Dillon, Joshua V and Murphy, Kevin},
  journal={arXiv preprint arXiv:1612.00410},
  year={2016}
}

@inproceedings{gao2024leveraging,
  title={Leveraging Large Language Models: Enhancing Retrieval-Augmented Generation with ScaNN and Gemma for Superior AI Response},
  author={Gao, Min and Lu, Peiqing and Zhao, Zihao and Bi, Xiaowei and Wang, Fa},
  booktitle={2024 5th International Conference on Machine Learning and Computer Application (ICMLCA)},
  pages={619--622},
  year={2024},
  organization={IEEE}
}

@inproceedings{long2024enhancing,
  title={Enhancing educational content matching using transformer models and infonce loss},
  author={Long, Yujian and Gu, Dian and Li, Xinrui and Lu, Peiqing and Cao, Jing},
  booktitle={2024 IEEE 7th International Conference on Information Systems and Computer Aided Education (ICISCAE)},
  pages={11--15},
  year={2024},
  organization={IEEE}
}

@inproceedings{tong2024integrated,
  title={An integrated machine learning and deep learning framework for credit card approval prediction},
  author={Tong, Kejian and Han, Zonglin and Shen, Yanxin and Long, Yujian and Wei, Yijing},
  booktitle={2024 IEEE 6th International Conference on Power, Intelligent Computing and Systems (ICPICS)},
  pages={853--858},
  year={2024},
  organization={IEEE}
}

@article{xue2026resilient,
  title={Resilient Routing: Risk-Aware Dynamic Routing in Smart Logistics via Spatiotemporal Graph Learning},
  author={Xue, Zhiming and Zhao, Sichen and Qi, Yalun and Zeng, Xianling and Yu, Zihan},
  journal={arXiv preprint arXiv:2601.13632},
  year={2026}
}

@article{wang2025reasoning,
  title={Reasoning or Retrieval? A Study of Answer Attribution on Large Reasoning Models},
  author={Wang, Yuhui and Li, Changjiang and Chen, Guangke and Liang, Jiacheng and Wang, Ting},
  journal={arXiv preprint arXiv:2509.24156},
  year={2025}
}

@article{wang2025self,
  title={Self-destructive language model},
  author={Wang, Yuhui and Zhu, Rongyi and Wang, Ting},
  journal={arXiv preprint arXiv:2505.12186},
  year={2025}
}

\end{document}